%% file: main.tex
\documentclass[journal,twoside,web]{ieeecolor}
\usepackage{tmi}
\usepackage{cite}
\usepackage{amsmath,amssymb,amsfonts}
\usepackage{graphicx}
\usepackage{textcomp}
\makeatletter
\let\NAT@parse\undefined
\makeatother
\usepackage[numbers,sort&compress]{natbib}
\usepackage{float}
\usepackage{stfloats}
\usepackage{amssymb}
\usepackage{bbding}
\usepackage{algorithm}
\usepackage{algpseudocode}
\usepackage{multirow}
\usepackage{longtable}
\usepackage{algorithmicx}
\usepackage{booktabs}
\algtext*{EndIf}
\algtext*{EndFor}
\usepackage{siunitx}
\usepackage{hyperref}
\usepackage{bm}
\usepackage{colortbl}
\usepackage{pifont}
\usepackage[margin=1in]{geometry}
\def\BibTeX{{\rm B\kern-.05em{\sc i\kern-.025em b}\kern-.08em
    T\kern-.1667em\lower.7ex\hbox{E}\kern-.125emX}}
\begin{document}

\title{High-Fidelity Medical Shape Generation via Skeletal Latent Diffusion}
\author{
Guoqing Zhang, Jingyun Yang, Siqi Chen, Anping Zhang, and Yang Li, \IEEEmembership{Member, IEEE}
\thanks{
This work was supported in part by the Natural Science Foundation of China (Grant 62371270) and the Major Key Project of PCL (Grant PCL2025A13, Pengcheng Laboratory).
Correspondence author: Yang Li. }
\thanks{Guoqing Zhang, Jingyun Yang, Siqi Chen, Anping Zhang are with Tsinghua Shenzhen International Graduate School, Tsinghua University, Shenzhen, China. Guoqing Zhang is also with Pengcheng Laboratory, Shenzhen, China. Yang Li is with School of AI, Chinese University of Hong Kong (Shenzhen). 
(email: zhanggq21@mails.tsinghua.edu.cn; 
yangl@cuhk.edu.cn).}
}
\maketitle

\begin{abstract}
Anatomy shape modeling is a fundamental problem in medical data analysis. However, the geometric complexity and topological variability of anatomical structures pose significant challenges to accurate anatomical shape generation. 
In this work, we propose a skeletal latent diffusion framework that explicitly incorporates structural priors for efficient and high-fidelity medical shape generation. We introduce a shape auto-encoder in which the encoder captures global geometric information through a differentiable skeletonization module and aggregates local surface features into shape latents, while the decoder predicts the corresponding implicit fields over sparsely sampled coordinates. New shapes are generated via a latent-space diffusion model, followed by neural implicit decoding and mesh extraction.
To address the limited availability of medical shape data, we construct a large-scale dataset, \textit{MedSDF}, comprising surface point clouds and corresponding signed distance fields across multiple anatomical categories. Extensive experiments on MedSDF and vessel datasets demonstrate that the proposed method achieves superior reconstruction and generation quality while maintaining a higher computational efficiency compared with existing approaches. Code is available at: \href{https://github.com/wlsdzyzl/meshage}{https://github.com/wlsdzyzl/meshage}.

\end{abstract}


\input{sections/introduction}
\input{sections/related_work}

\input{sections/method}
\input{sections/experiment}

\input{sections/discussion}

\bibliographystyle{IEEEtranN}
\bibliography{tmi}

\end{document}

%% file: sections/introduction.tex
\begin{figure*}[h]
	\centering
	\includegraphics[width=\linewidth, angle=0]{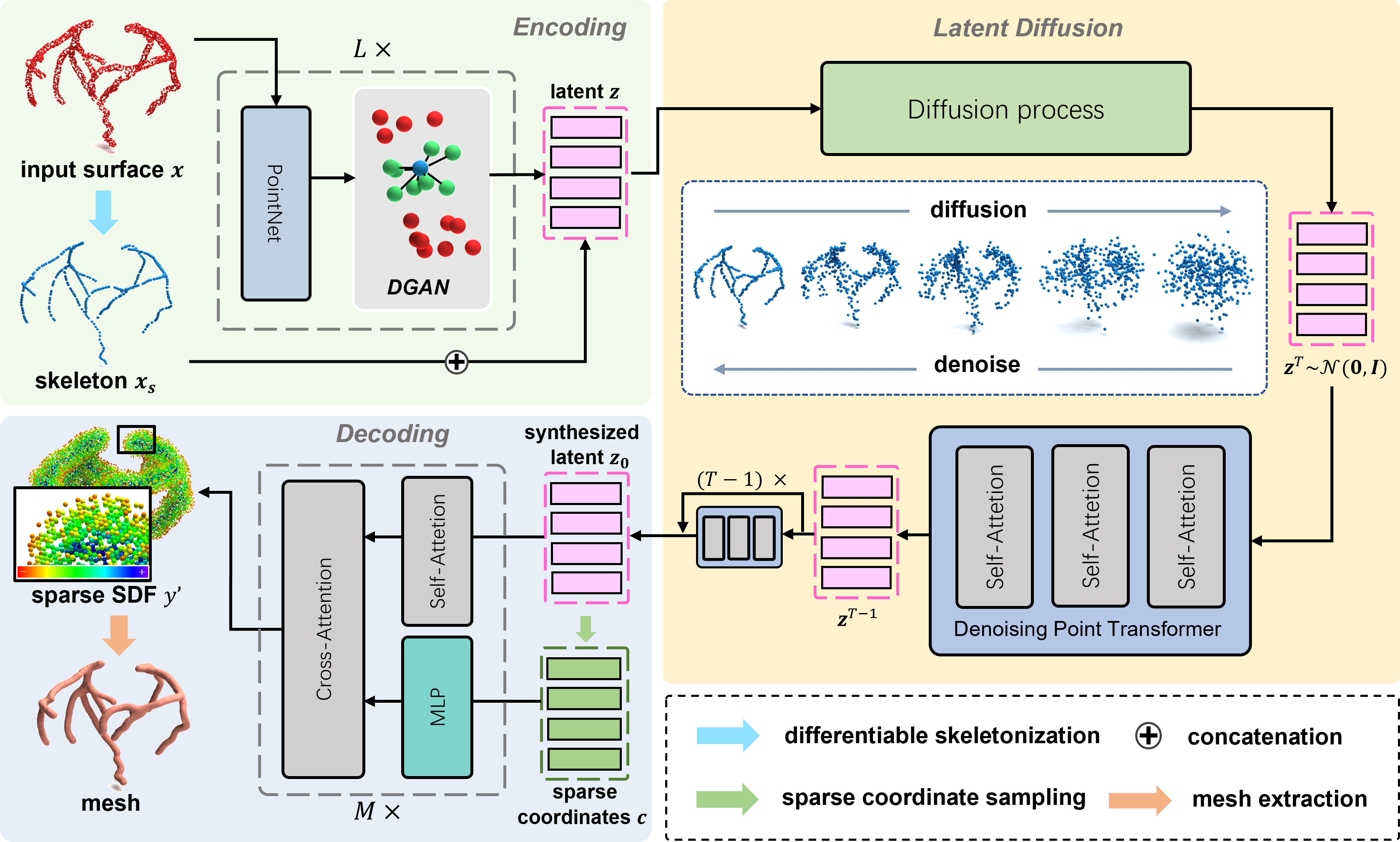}
        \vspace{-1em}
	\caption{A semantic overview of the proposed architecture. We first train a VAE where the encoder maps the input surface and online-computed skeleton to a latent point set, which serves as a shape signature for the decoder to predict SDF values at queried spatial coordinates. A diffusion model is then trained in the latent feature space. For shape synthesis, new latents are sampled via the reverse diffusion process and decoded into neural implicit fields, which are subsequently converted into 3D shapes.}
	\label{fig:overview}
 \vspace{-1.0em}
\end{figure*}
\section{Introduction}
\label{sec:intro}
Medical shape generation has significant potential for a wide range of applications, including surgical planning and simulation, statistical anatomical modeling, and medical education. However, it remains a challenging problem due to the intricate geometry, structural variability, and complex topology inherent in anatomical structures. These difficulties are further compounded by the scarcity of large-scale medical shape datasets, resulting from limited data availability, privacy constraints, and the high cost of expert annotations.

Recently, diffusion models have emerged as a powerful framework for 3D shape and point cloud synthesis. Point cloud diffusion methods \cite{luo2021diffusion,zhou20213d,zeng2022lion} operate directly in Euclidean space by iteratively denoising noisy point sets to recover surface points. Although conceptually straightforward, these models often struggle to converge on medical point clouds due to the complex anatomical geometries and thin tubular structures. Over the past few years, researchers have explored structure-aware generative models. \citet{Sin_TrIND_MICCAI2024} perform diffusion on hierarchical tree representations, while graph-based approaches \cite{Pra_3D_MICCAI2024,CheSiq_Hierarchical_MICCAI2025, guo2025vesseldiffusion} model tubular structures as node-edge graphs. These representations are limited to tree- or graph-like geometries and generalize poorly to complex anatomical shapes with surface variability. \citet{petrov2024gem3d} develop GeM3D that uses the medial axis as a geometric prior for shape synthesis. Nonetheless, it relies on pre-computed skeletons extracted through a non-differentiable process, preventing flexible integration with the learning framework. Moreover, these methods focus only on global structure and fail to capture fine-grained local geometric details, which are crucial for accurate anatomical modeling.

To address these challenges, we propose a skeleton-guided medical shape generation framework that performs diffusion in a learned latent space. Since point clouds are sparsely sampled and not directly suitable for downstream applications such as surgical simulation, we adopt an implicit shape auto-encoder that maps point sets to continuous signed distance fields (SDF). 
Our model introduces a differentiable skeletonization module to extract skeletal representations, which capture the global geometry and topology of anatomical shapes and provide a compact structural abstraction for aggregating local surface features in the encoder. Compared with surface representations, skeletons offer a significantly more compact description of shape, making them easier to model and generate while also serving as explicit structural priors for subsequent shape decoding. Furthermore, we leverage skeletons to guide sparse coordinate sampling during inference, focusing computation on regions near the underlying shape and thereby accelerating SDF prediction. The differentiable formulation enables seamless integration of skeletonization into networks for end-to-end feature learning. 
To generate a new shape, skeletal latents are synthesized using a Transformer-based point diffusion model~\cite{Karras2022edm} trained in the latent space. The synthesized latents are then converted to triangle meshes through SDF decoding and mesh extraction. The technical contributions of this work are listed as follows:

\vspace{0.05em}
\begin{itemize}
    \item We propose a novel generative framework that performs diffusion in a compact structure-aware latent space for medical shapes.
    \item We introduce a shape auto-encoder that jointly encodes global structure and local surface details into skeletal latent representations, enabling efficient neural implicit field prediction.
    \item We build MedSDF, a large-scale multi-category medical shape dataset that provides paired surface point clouds and SDF volumes for training neural implicit models and performance evaluation.
    \item We conduct extensive experiments on MedSDF and vessel datasets, demonstrating the effectiveness and efficiency of the proposed method for medical shape reconstruction and generation. 
\end{itemize}

%% file: sections/related_work.tex
\section{Related Works}
\subsection{Skeleton Representation}
The computation of medial skeletons from surfaces is a well-studied problem in geometry processing. Traditional skeletonization methods mainly rely on analytic approximations, including Voronoi-based approaches \cite{10.1145/280814.280947, 10.1145/3197517.3201396}, Power diagrams \cite{10.1145/3550454.3555465}, and morphological operations such as surface contraction \cite{10.1145/2461912.2461913, 10.1111/j.1467-8659.2012.03178.x}. Such methods are typically non-differentiable, computationally expensive, and often restricted to specific input formats (e.g., meshes). For a comprehensive overview of geometric approaches to skeletonization, we refer readers to the survey by \citet{tagliasacchi20163d}. More recently, learning-based methods \cite{yin2018p2pnet, 10.1016/j.cag.2023.07.020, Lin_2021_CVPR} have been developed to learn skeletal representations from point clouds. These models struggle to maintain correct topology and introduce extra parameters that increase the end-to-end training overhead. 

\subsection{3D Generation}
Diffusion models \cite{10.5555/3495724.3496298} have emerged as a powerful framework for 3D shape generation. Point cloud diffusion methods \cite{luo2021diffusion,zhou20213d, zeng2022lion} operate directly in Euclidean space by iteratively denoising noisy point sets to recover surface geometry. Neural implicit generative models \cite{Park_2019_CVPR,pmlr-v80-achlioptas18a,cheng2022sdfusion,10.1145/3635304,10.1145/3592442} instead parameterize SDF or occupancy with neural networks, enabling high-resolution surface generation from latent embeddings. However, these approaches often produce shapes with suboptimal geometric fidelity and structural artifacts. The method most closely related to ours is GeM3D \cite{petrov2024gem3d}, which employs pre-computed medial axes for shape synthesis from neural fields. Nevertheless, it requires a large number of skeletal points to adequately capture global structure and exhibits limited capacity to generate fine-grained local geometry. Moreover, estimating neural implicit fields at high resolution incurs substantial computational time. Several other structure-aware approaches have also been specially designed for vascular modeling by representing tubular shapes as hierarchical implicit fields \cite{Sin_TrIND_MICCAI2024} or vessel graphs \cite{Pra_3D_MICCAI2024,CheSiq_Hierarchical_MICCAI2025,guo2025vesseldiffusion}. Such methods are inherently restricted to tree- or graph-like topologies and are difficult to generalize to other anatomical surfaces.

%% file: sections/method.tex
\section{Methodology}
The proposed generative framework involves a two-stage training process. Firstly, we develop an effective auto-encoder to map the input point sets into a skeletal latent space. Then, we employ a diffusion model in the latent space for shape generation. In the following section, we provide an overview of the overall network architecture, followed by a detailed description of the encoding-decoding pipeline and shape latent synthesis.
\begin{figure*}[!htbp]
	\centering
	\includegraphics[width=\linewidth, angle=0]{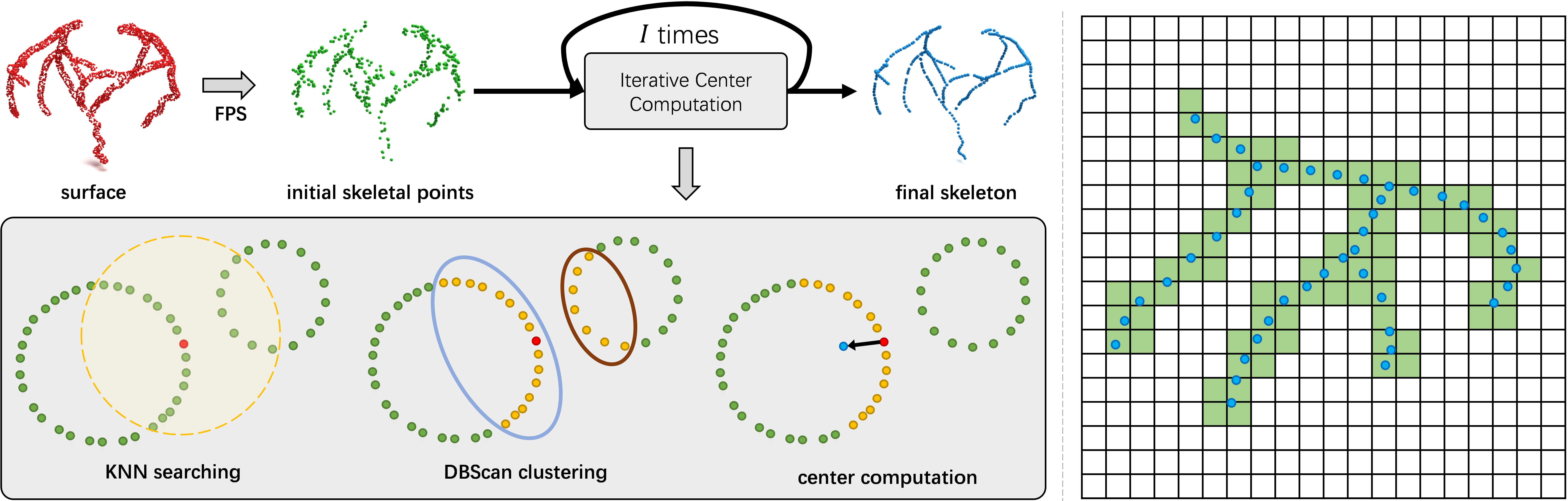}
        \vspace{-1em}
	\caption{Left: differentiable skeletonization. Right: skeleton-guided coordinate sampling (right); only $10\%$ voxels (green grids) near the skeletal points (blue dots) are sampled for SDF calculation during inference. }
	\label{fig:module}
 \vspace{-1em}
\end{figure*}
\subsection{Network Architecture}
An overview of our framework is presented in Fig.~\ref{fig:overview}. In the first training stage, an input training batch consists of a point set, a query coordinate set randomly sampled from the truncated SDF volume, and the corresponding SDF values. Given a point set, we first extract its skeleton using a differentiable geometric algorithm. The encoder then embeds surface and skeletal information into a shape latent representation, while the decoder learns a deep implicit function that predicts signed distances to the surface from the latent. In the second stage, we train a diffusion model on the learned latent representations. The generated shape latents are subsequently decoded to SDF volumes and converted to dense meshes using the Marching Cubes algorithm~\cite{10.1145/37401.37422}.

\subsection{Shape Auto-Encoder}
\subsubsection{Shape Encoding}
To encode a shape, we first extract its skeleton $\mathbf{x}_s$ from the input point cloud $\mathbf{x}$ and employ a dual-branch point cloud encoder to compute the shape latent representation $\mathbf{z}$. Similar to~\cite{zhang2025lagem}, we apply a learnable standardization on the latent code to encourage a well-structured latent space. Furthermore, since the skeleton provides a compact structural representation of the original shape, we explicitly incorporate the skeletal points into the latent representation. As will be discussed later, this explicit structural prior improves both the efficiency and effectiveness of shape decoding.
The encoding process can be formulated as:
$$
\mathbf{z} = \operatorname{Cat}\left(\mathbf{x}_s, \operatorname{Std}(\mathcal{E} (\mathbf{x, x_s}))\right ),
$$
where $\mathbf{x} \in \mathbb{R}^{n \times 3}$, $\mathbf{x}_s \in \mathbb{R}^{n_s \times 4}$, and $\mathbf{z} \in \mathbb{R}^{n_s \times (4+d)}$. Here, $\mathcal{E}$ denotes the encoder, $\operatorname{Cat}(\cdot, \cdot)$ denotes concatenation, and $\operatorname{Std}(\cdot)$ represents the standardization operation implemented via layer normalization, skeleton $\mathbf{x}_s$ is composed of three coordinate channels and one additional channel representing the radius. The variables $n$, $n_s$, and $d$ denote the numbers of surface points, skeletal points, and latent channels, respectively. 

\vspace{0.5em}
\textbf{Differentiable Skeletonization:} The left part of Fig.~\ref{fig:module} illustrates our geometric skeletonization module. We first apply farthest point sampling (FPS) on the surface to select $n_s$ initial skeletal points, which are refined through $I$ iterations of center computation to obtain the final skeleton. In each iteration, a K-nearest neighbor (KNN) search identifies local surface neighborhoods for each skeletal point. Since KNN may retrieve points from different local structures, DBSCAN clustering~\cite{10.5555/3001460.3001507} is applied to separate geometric components, and the center of the cluster associated with the current skeletal point is used to update its position. By repeating this process, the skeletal points progressively converge to the final skeleton representation. Mathematically, the proposed skeleton extraction can be written as:
$$
\operatorname{Ske}(\mathbf{x}, n_s, i) = \{x^i_{s,1}, x^i_{s,2}, ..., x^i_{s, n_s}\},~ 1 < i \le I,
$$
where
$$
x_{s, j}^{i} = \frac{1}{\vert\mathcal{C}\left(x_{s,j}^{i-1}\right)\vert}\sum_{p \in \mathcal{C}\left(x_{s,j}^{i-1}\right)}p,~ 1 \le j  \le n_s.
$$
Here $\operatorname{Ske}(\mathbf{x}, n_s, i)$ computes $n_s$ skeletal points of $\mathbf{x}$ at iteration $i$, and $\mathcal{C}\left(x_{s,j}^{i-1}\right)$ denotes the clustered neighbors of $x_{s,j}^{i-1}$ from $\mathbf{x}$. The iteration number $I$ and neighborhood size $K$ control the skeletonization strength. In our implementation, we set $I=2$ and $K = 32$, to allow an accurate skeleton extraction for tubular structures while preserving the overall morphology of general shapes. Furthermore, we adopt a hierarchical refinement strategy in which We first compute $4n_s$ intermediate skeletal points $\hat{\mathbf{x}}_s$ from surface point $\mathbf{x}$ and then extract $n_s$ final skeletal points $\mathbf{x}_s$ from $\hat{\mathbf{x}} _s$:
$$
\mathbf{x}_s = \operatorname{Ske}(\operatorname{Ske}(\mathbf{ x }, 4n_s, I), n_s, I).
$$
Since each update is defined as the center of a subset of surface points, the entire process is differentiable and can be integrated into point networks for end-to-end training.
\vspace{0.5em}
\textbf{Surface Feature Aggregation:} 
While skeletons effectively capture global shape structure, local geometric details may be lost during extraction. To address this, we introduce a dual-branch encoder that integrates surface information into the skeletal latent. The surface points $\mathbf{x}$ and skeletal points $\mathbf{x}_s$ are first mapped into feature space using two MLPs, producing initial surface features $\mathbf{f}^0$ and skeleton features $\mathbf{f}_s^0$. We then construct an $L$-level feature aggregation network to progressively refine both surface and skeletal features. At the $l$-th level, a shared PointNet~\cite{qi2017pointnet} layer extracts deeper features $\mathbf{f}^l$ and $\hat{\mathbf{f}}_s^l$. A dynamic graph aggregation network (DGAN) further aggregates local surface information by performing KNN search in feature space and grouping $k_l$ neighbors of $\hat{\mathbf{f}}_s^l \in \mathbb{R}^{n_s\times d_l}$ from $\mathbf{f}^l \in \mathbb{R}^{n\times d_l}$, producing grouped features $\mathbf{f}_g^l \in \mathbb{R}^{n_s \times k_l \times d_l}$, where $d_l$ denotes the feature dimension at level $l$. Finally, max pooling is applied along the neighbor dimension to obtain the $l$-th level skeletal features $\mathbf{f}_s^l \in \mathbb{R}^{n_s \times d_l}$. This operation is conceptually similar to PointNet++~\cite{qi2017pointnet++}, except that we adopt KNN-based grouping in feature space rather than ball-query-based grouping in Euclidean point space. The feature updates can be formulated as:
$$
\mathbf{f}^l = \mathcal{P}(\mathbf{f}^{l-1}),~ \mathbf{f}_s^l = \operatorname{Max}_g(\operatorname{QueryGroup}(\mathcal{P}(\mathbf{f_s}^{l-1}), \mathbf{f_l})),
$$
where $\mathcal{P}(\cdot)$ denotes the PointNet-based feature transformation, $\operatorname{QueryGroup}(\cdot, \cdot)$ performs KNN-based grouping and feature aggregation, and $\operatorname{Max}_g(\cdot)$ denotes max pooling along the neighbor dimension.

\vspace{0.5em}
\subsubsection{Shape Decoding}
Our decoder takes the latent code $\mathbf{z}$ as a shape signature and predicts the SDF values $y'$ at query coordinates $\mathbf{q}$. Before decoding, we apply a de-standardization to map latent points back to the feature space. The decoding process is defined as
$$
y' = \mathcal{D}\left(\mathbf{q}, \operatorname{Std}^{-1}(\mathbf{z})\right),
$$
where $\mathbf{q} \in \mathbb{R}^{m\times3}$ denotes $m$ query coordinates and $\mathcal{D}$ is the decoder. $\operatorname{Std}^{-1}(\cdot)$ denotes the inverse standardization implemented via learnable scaling and shifting, applied only to the last $d$ channels of $\mathbf{z}$. The first four channels contain the skeleton coordinates and radius, which provide an explicit geometric prior for the decoder. 

\vspace{0.5em}
\textbf{Neural Implicit Function:} 
In the decoding stage, we model the neural implicit function as a learnable network that decodes shape information from latent points $\mathbf{z}$ and predicts the corresponding signed distance values $\mathbf{y}' \in \mathbb{R}^m$ for query coordinates $\mathbf{q}$.
Before being processed, the query coordinates are encoded using sinusoidal positional embeddings~\cite{vaswani2017attention} to preserve spatial information. The latent points are fed into a Transformer block to further interpret the underlying shape information, yielding latent feature $\mathbf{f}_z \in \mathbb{R}^{n_z\times d}$, while the query coordinates are processed by an MLP to obtain coordinate features $\hat{\mathbf{f}}_q \in \mathbb{R}^{m\times d}$. This design is motivated by the fact that query coordinates might be arbitrarily sampled, and self-attention (SA) among them provides limited semantic meaning. Subsequently, cross-attention (CA) is performed between $\mathbf{f}_z$ and the $\hat{\mathbf{f}}_q$ to obtain the coordinate feature $\mathbf{f}_q$. Formally, we define the feature fusion procedure as:
$$
\mathbf{f}_q \gets \operatorname{CA}(\operatorname{MLP}(\mathbf{f}_q), \operatorname{SA}(\mathbf{f}_z)).
$$
After $M$ feature fusion steps, the resulting coordinate feature is projected through a linear transformation to predict the SDF values.

\vspace{0.5em}
\textbf{Sparse Coordinate Sampling:} 
A cubic volume with resolution $100^3$ contains one million voxels, making full SDF estimation computationally expensive. However, only sparse voxels near the surface contain most of the geometric information, which motivates the truncated signed distance function (TSDF) technique \cite{6162880}. Following this idea, we randomly sample $m$ coordinates from the truncated SDF volume during training. During inference, since the surface locations and SDF volume are unknown, we adopt skeleton-guided coordinate sampling as illustrated in the right part of Fig.~\ref{fig:module}. Skeleton points are first extracted from the shape latent representation, and the top $p$ fraction of voxels closest to the skeleton are selected for SDF prediction, while the remaining voxels are assigned a value of $-1$. Empirically, setting $p\in[0.1,0.3]$ provides a good trade-off between efficiency and shape quality of shape decoding.

\begin{figure*}[!htbp]
	\centering
	\includegraphics[width=\linewidth, angle=0]{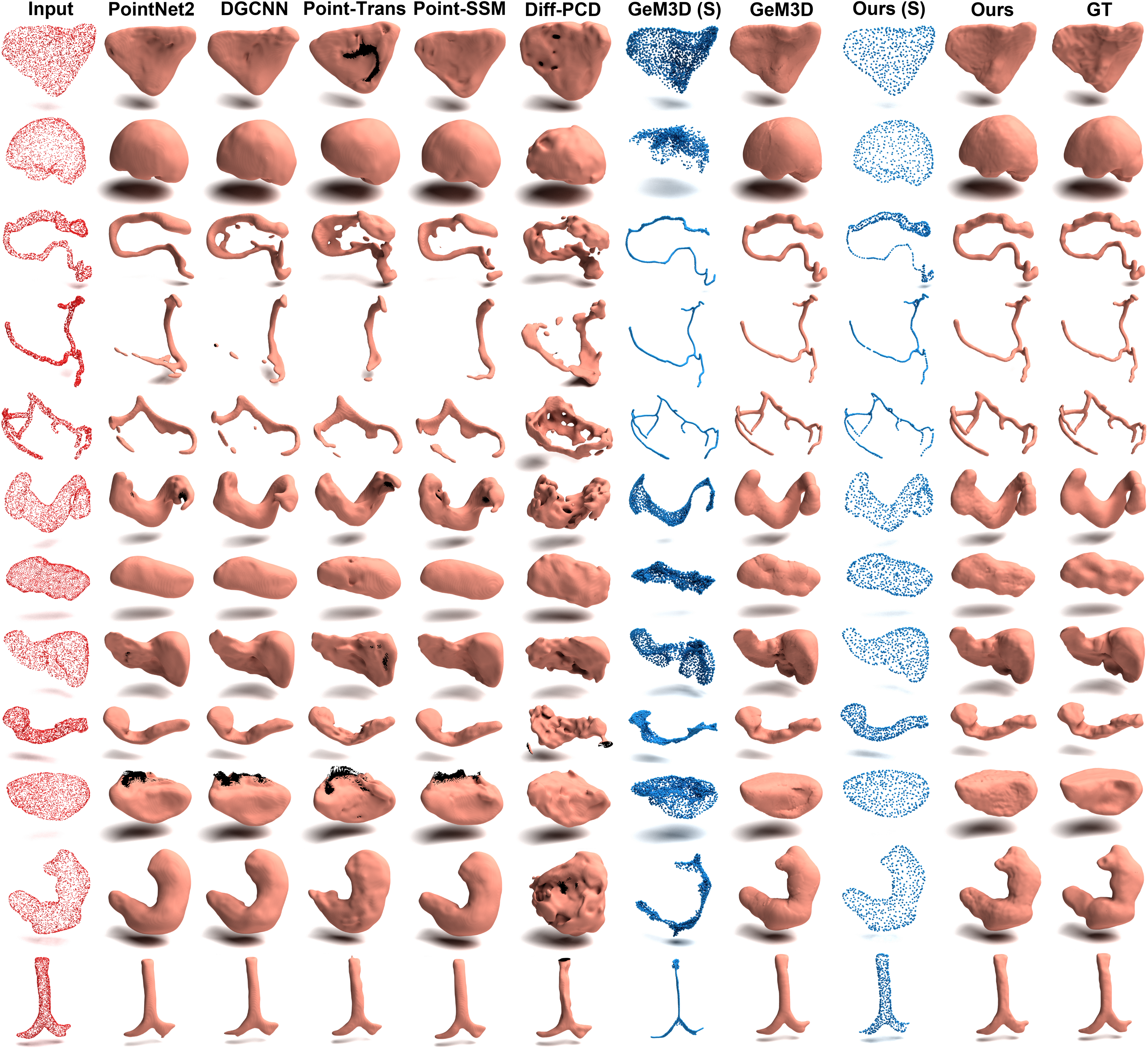}
        \vspace{-1.5em}
	\caption{Qualitative comparison of shape reconstruction on MedSDF dataset. From top to bottom, the rows correspond to samples of the bladder, brain, colon, right coronary artery (RCA), left coronary artery (LCA), duodenum, gallbladder, liver, pancreas, spleen, stomach, and trachea. \textit{\textbf{GeM3D (S)}} and \textit{\textbf{Ours (S)}} denotes the skeletal points computed by GeM3D and our method, respectively.}
	\label{fig:recon}
 \vspace{-1em}
\end{figure*}

\vspace{0.5em}
\subsubsection{Loss Function}
Our encoder and decoder are optimized by minimizing the mean squared error (MSE) between the predicted and ground-truth SDF values. We propose an extra skeleton-based constraint, according to the fact that the distance between a skeletal point and the closest surface is the radius. Overall, our loss function can be written as:
$$
\mathcal{L} = \operatorname{MSE}(y', y) + \lambda \operatorname{MSE}\left(\mathcal{D}(\mathbf{x}_s[\text{p}]), \operatorname{Std}^{-1}(\mathbf{z})), \mathbf{x}_s[\text{r}]\right),
$$
where $\mathbf{x}_s[\text{p}]$ and $\mathbf{x}_s[\text{r}]$ extract the position and radius of the skeletal points $\mathbf{x}_s$, $\lambda$ is the strength of skeleton constraints and empirically set as $1.0$. 

\subsection{Shape Latent Generation}
Inspired by \cite{rombach2021highresolution,10.1145/3592442}, we propose a skeletal latent diffusion framework. Specifically, we initialize $n_s$ points from Gaussian noise $\mathbf{z}_{T} \sim \mathcal{N}(\mathbf{0}, \sigma_{T}^2 \mathbf{I})$, where $T$ and $\sigma_{T}$ denote the maximum sampling steps and the highest noise level, respectively. Generation is performed by solving a probability flow ordinary differential equation (ODE) \cite{Karras2022edm} over a sequence of time steps, progressively transforming noise samples into structured skeletal latent points:
$$
\text{d}\mathbf{z}_t = -\frac{\partial \sigma_t}{\partial t}\sigma_t\operatorname{Score}_\theta(\mathbf{z}_t, \sigma_t)\, \text{d}t,
$$
where $\sigma_t$ defines the continuous noise schedule, and $\operatorname{Score}_\theta(\cdot, \cdot)$ denotes the learned score function parameterized as a Transformer-based point network $\theta$. This formulation enables deterministic sampling while maintaining consistency with the underlying stochastic diffusion process. To support multi-category generation, we further adopt classifier-free guidance \cite{ho2021classifierfree}. The guided score function is defined as:
$$
\operatorname{Score}_{\theta}(\mathbf{z}_t, \sigma_t, c) = (1+w)\theta(\mathbf{z}_t,  \sigma_t, \varnothing) -w \theta(\mathbf{z}_t,  \sigma_t, c),
$$
where $c$ denotes the category condition and $w$ is the guidance weight. Since the number of skeletal latent points is typically an order of magnitude smaller than the number of surface points, the proposed approach enables faster sampling compared to point-based diffusion models.

%% file: sections/experiment.tex
\section{Experiments}

\subsection{Datasets}
To validate our method, we introduce \textbf{MedSDF}, a large-scale medical shape dataset containing 12,472 samples across 14 anatomical categories for category-conditioned shape reconstruction and generation. Each sample includes a point cloud, an SDF volume of $100^3$, and a reconstructed surface mesh. We process over 10,000 samples from 12 anatomical categories in MedShapeNet~\cite{li2024medshapenet} and compute their corresponding SDF volumes. To increase morphological diversity, we additionally incorporate vessel data from ImageCAS~\cite{ZENG2023102287}, where each coronary artery is segmented into left and right branches. The dataset is randomly split into 5 folds, with 90\% and 10\% of samples from the first four folds used for training and validation, respectively, and the remaining fold reserved for testing.

We further conduct experiments on two vascular datasets to assess performance on tubular shapes. CoW~\cite{Junayed_2024} contains 279 MRI scans of intracranial artery segmentation, while ImageCAS~\cite{ZENG2023102287} includes 1000 CT images of coronary artery segmentations. For these datasets, we adopt a 5-fold scheme: the first three folds are used for training, the fourth for validation, and the fifth for testing.

\begin{figure*}[!htbp]
	\centering
	\includegraphics[width=\linewidth, angle=0]{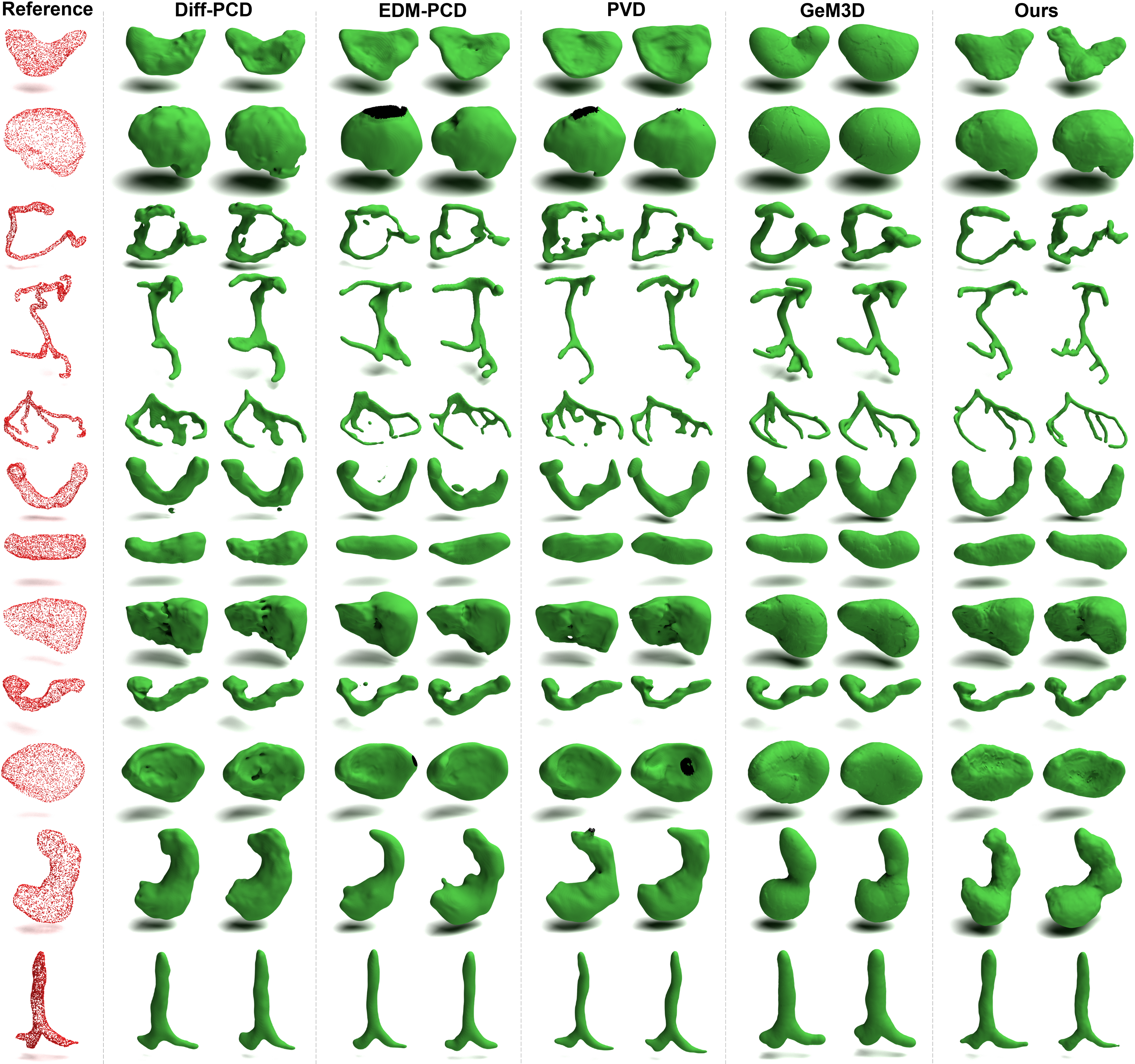}
        \vspace{-1.5em}
	\caption{Qualitative comparison of shape generation on MedSDF dataset. From top to bottom, the rows correspond to samples of the bladder, brain, colon, right coronary artery (RCA), left coronary artery (LCA), duodenum, gallbladder, liver, pancreas, spleen, stomach, and trachea. For each category and method, we generate 500 samples and display the two nearest neighbors of the reference point cloud.}
	\label{fig:gen}
 \vspace{-1em}
\end{figure*}
\input{tables/recon}
\subsection{Evaluation Metrics}
Following previous works \cite{luo2021diffusion, zhou20213d, petrov2024gem3d}, we use Chamfer Distance (CD) and Earth Mover's Distance (EMD) to quantify the overall fidelity of shape reconstruction. To further assess the precise match between the predicted and ground-truth shapes, we also report the Hausdorff Distance (HD) and the F1-Score (F1), calculating the latter with a distance threshold of $0.06$. For reconstruction evaluation, input point clouds are sub-sampled to 2,560 points via FPS to better preserve fine structures.

For shape generation, we use the Fréchet Inception Distance (FID) and Kernel Inception Distance (KID) as primary metrics, which measure distribution differences in feature space. The point cloud features are extracted using a DGCNN~\cite{wang2019dynamic} network pre-trained on the MedPointS~\cite{zhang2024flemme} dataset. We further report Minimum Matching Distance (MMD), Coverage (COV), and 1-Nearest Neighbor Accuracy (1-NNA), computed with both CD and EMD. MMD measures the average distance from each generated sample to its nearest neighbor in the test set, and COV evaluates the coverage of generated samples over the true distribution. 1-NNA reports the accuracy of a 1-NN classifier distinguishing generated and real samples; a score close to 50\% suggests the two distributions are nearly indistinguishable. For all generation metrics, input point clouds are sub-sampled to 2,048 points via FPS.

\subsection{Experimental Setup}
\subsubsection{Baseline Models} 
We evaluate the reconstruction performance of the proposed auto-encoder by comparing it with state-of-the-art point cloud encoders, including PointNet++~\cite{qi2017pointnet++}, DGCNN~\cite{wang2019dynamic}, PCT~\cite{guo2021pct}, and Point-SSM~\cite{ZhaGuo_Hierarchical_MICCAI2025}. These methods represent strong baselines for geometric feature learning from point clouds. For fair comparison, all methods adopt the FoldingNet decoder~\cite{yang2018foldingnet}. We also compare with Diff-PCD~\cite{luo2021diffusion} and GeM3D~\cite{petrov2024gem3d}, two diffusion-based generative models. Diff-PCD performs point cloud auto-encoding conditioned on a PointNet-based shape embedding~\cite{qi2017pointnet}, while GeM3D predicts skeletons using P2P-Net~\cite{yin2018p2pnet} and reconstructs shapes through an implicit-field auto-encoder.

For shape generation, we compare with Diff-PCD~\cite{luo2021diffusion}, EDM-PCD, PVD~\cite{zhou20213d}, and GeM3D~\cite{petrov2024gem3d}. Diff-PCD and EDM-PCD perform diffusion directly in point space, with the latter adopting the EDM framework with a Transformer backbone~\cite{Karras2022edm}. PVD combines voxel convolution and attention for noise prediction, while GeM3D employs two separate diffusion models for skeleton and latent prediction.
\input{tables/gen}
To evaluate the performance of our method on vascular shape reconstruction and generation, we select Diff-Vessel \cite{guo2025vesseldiffusion} as a baseline, which is the-state-of-the-art vessel generation framework. Diff-Vessel extracts graphs on 2D vessel masks as a guide for point cloud prediction. For vessel reconstruction, segmentation masks from the test set are used as conditions. For vessel generation, an image generative model \cite{10.5555/3495724.3496298} is trained on each vessel dataset to synthesize 2D masks.

\input{tables/vessel}
\subsubsection{Implementation Details}
For MedSDF, the input point cloud is sub-sampled to 2,560 points using FPS, and the number of latent points is set to 256. During training, 2,560 voxel coordinates with ground-truth SDF values are sampled, with 90\% drawn from the truncated region (truncation value 0.1) and the remaining 10\% from outside to provide coarse supervision for distant regions and improve scale generalization; during inference, 30\% of voxels are sampled using the proposed skeleton-guided strategy for mesh extraction. For vessel datasets, the input and latent point sizes are 4,096 and 400, respectively. We sample 4,096 SDF coordinates during training, all from the truncated region since the task involves single-category reconstruction, and sample 10\% of voxels during inference due to the thin and elongated nature of vascular structures. For Diff-PCD, GeM3D, and Diff-Vessel, we use the authors’ official implementations, while other baselines are implemented using the medical image learning platform~\cite{zhang2024flemme}. Reconstruction models are trained for 150 epochs with an initial learning rate of $10^{-4}$ and a batch size of 6. Diff-PCD is trained for 1,000 epochs with an initial learning rate of $10^{-3}$ and a batch size of 96, with separate models trained per category due to the lack of multi-class conditioning. PVD and EDM-PCD are trained for 100 epochs with a batch size of 8. Since our diffusion operates in a compact latent space, it requires substantially less GPU memory, enabling training with a batch size of 128 for 1,000 epochs. GeM3D uses a latent size of 2,048 and follows its default setting of 1,000 epochs with a batch size of 8. All models are trained on an A800 80GB GPU with a linearly decaying learning rate schedule.

\subsection{Results}
\subsubsection{Quantitative and Qualitative Analysis} Tab.~\ref{tab:recon} shows the quantitative results for conditional medical shape reconstruction. The measures are averaged over 14 categories of MedSDF. For all reconstruction metrics, our method outperforms other competing works, demonstrating a robust capability for medical shape understanding. Fig.~\ref{fig:recon} is a qualitative visualization of the reconstruction results, in which we also display the computed skeletons by GeM3D and ours. Our method captures richer surface details of general medical shapes using significantly fewer skeletal points, which also leads to a more compact representation that is easier to model and more computationally efficient. 
\begin{figure}[!htbp]
	\centering
	\includegraphics[width=\columnwidth, angle=0]{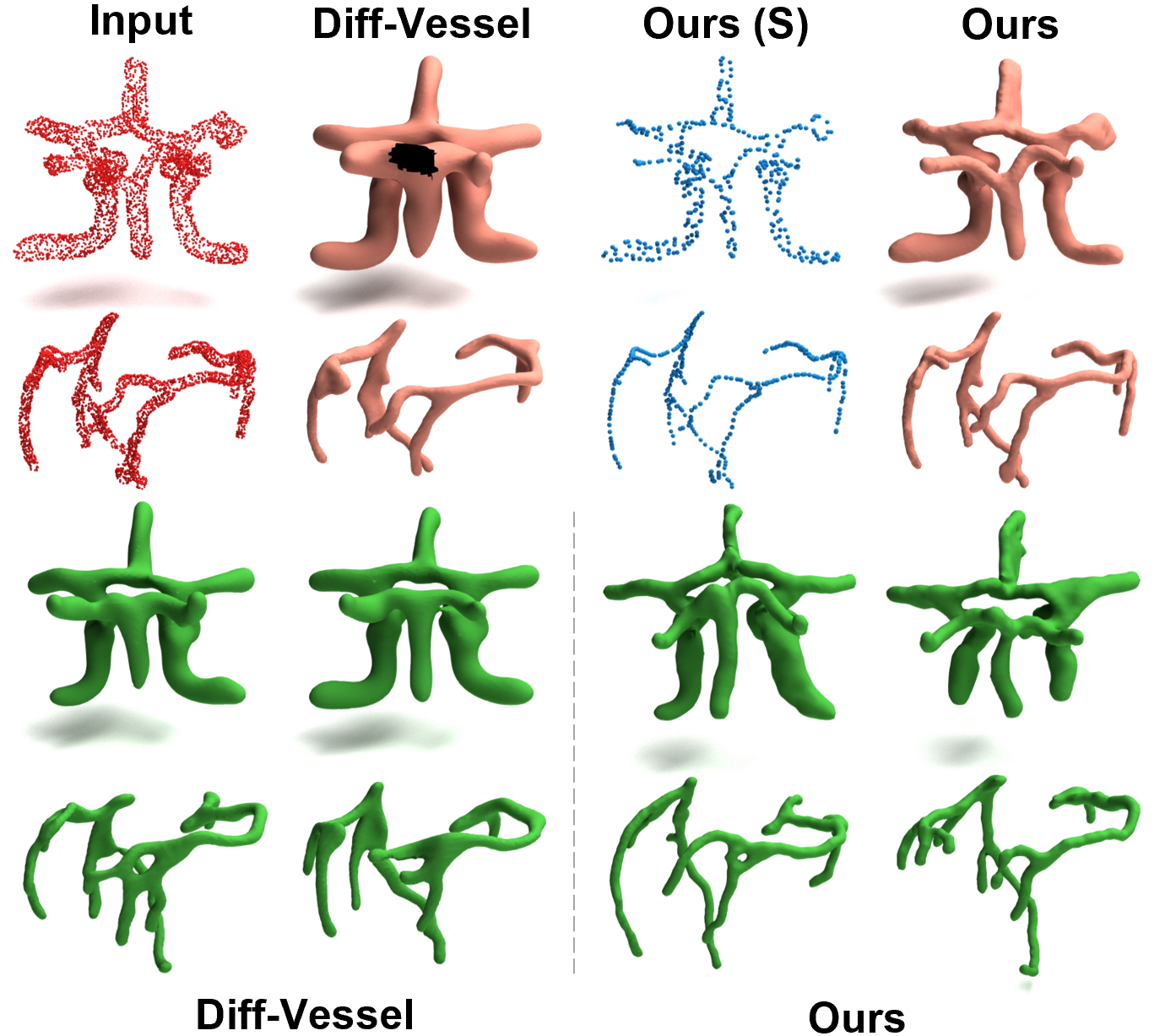}
        \vspace{-1em}
	\caption{Qualitative comparison of tubular shape reconstruction and generation. The top two rows correspond to reconstruction results of input point clouds from CoW and ImageCAS datasets, while the bottom two rows display the neighbors of input point clouds among the generated shapes. }
	\label{fig:vessel}
\end{figure}
We report the evaluation of generation quality in Tab.~\ref{tab:gen}. Our method consistently ranks among the top two across all metrics and achieves the best performance on seven of these measures. We observe that, despite achieving comparable reconstruction accuracy, GeM3D exhibits limited generative capability. This may be attributed to the insufficient aggregation of neighborhood information in its skeleton representation, as well as the absence of explicit latent-space regularization in the auto-decoder. Diff-PCD achieves the best MMD-CD score; however, as discussed by \cite{9010395}, MMD is insensitive to low-quality point clouds. The lower scores of Diff-PCD in FID and KID suggest inferior generation quality. On the other hand, both EDM-PCD and PVD generate high-fidelity shapes; their poor performance on the COV and NNA metrics indicates lower diversity in the generated outputs. The generation time per sample (GTS) is also reported. Our method achieves a favorable balance, producing high-quality and diverse shapes while maintaining efficient generation. Qualitative comparisons in Fig.~\ref{fig:gen} further show that our method synthesizes shapes with coherent global structure and richer surface details.

The experimental results of reconstruction and generation on vessel datasets are summarized in Tab.~\ref{tab:vessel}. Diff-Vessel guides the diffusion process using a single-view 2D mask, which may provide insufficient constraints for accurate shape reconstruction. In contrast, the proposed method demonstrates a significantly enhanced capability for reconstructing complex vascular geometries while maintaining comparable performance in the generation task. A qualitative visual comparison of these results is shown in Fig.~\ref{fig:vessel}.

\input{tables/ablation}
\subsubsection{Ablation Studies}
We conduct ablation studies on CoW, ImageCAS, and MedSDF datasets to evaluate the contribution of individual modules as reported in Tab.~\ref{tab:ablation}, including differentiable skeletonization (DS), skeleton constraints (SC), and latent attention (LA). We first construct a baseline model using skeletons pre-computed by GeM3D~\cite{petrov2024gem3d}, and then replace them with DS. As illustrated in Fig.~\ref{fig:recon}, the offline skeleton fails to adequately capture the morphology of non-tubular shapes, resulting in poor reconstruction and generation performance on CoW and MedSDF datasets. In contrast, our method extracts a more compact representation that generalizes better to shapes with diverse geometries. In addition, incorporating SC consistently improves both reconstruction and generation performance. Combining with LA, the full model achieves the best overall performance across all datasets.

%% file: tables/recon.tex
\begin{table}[h]
\caption{Shape reconstruction evaluation on the MedSDF dataset. CD is multiplied by $10^2$, EMD is multiplied by $10$, and HD is multiplied by $10$. The best results are denoted in \textbf{Bold}.}
\label{tab:recon}
        \centering
        
        \footnotesize
            \begin{tabular}{c|cccc}
                \toprule
                Method & CD ($\downarrow$)  & EMD ($\downarrow$) & HD ($\downarrow$) & F1 ($\%, \uparrow$)\\
                \midrule
                PointNet++ & 3.811 & 1.849 & 2.045 & 85.97  \\
                DGCNN & 3.681 & 1.793 & 2.104 & 87.29 \\ 
                PCT & 5.577 & 3.875 & 3.029 & 67.98 \\
                Point-SSM & 3.876 & 2.012 & 2.336 & 85.30 \\ 
                Diff-PCD & 6.001 & 1.278 & 3.023 & 61.47 \\ 
                GeM3D & 2.672 & 0.643 & 1.260 & 97.86 \\
                Ours & \textbf{2.314} & \textbf{0.447} & \textbf{1.136} & \textbf{98.24} \\
                 \bottomrule
                \end{tabular} 
\end{table}
                

%% file: tables/gen.tex
        

\begin{table}[h]
\caption{Category-conditioned shape generation evaluation on the MedSDF dataset. KID is multiplied by $10$, MMD-CD is multiplied by $10^2$, and MMD-EMD is multiplied by $10$. The best results are denoted in \textbf{Bold}. The second-best score is \underline{underlined}.}
\label{tab:gen}
        \centering
        
        \footnotesize
     \resizebox{\columnwidth}{!}{%
            \begin{tabular}{r|ccccc}
                \toprule
                Metric & Diff-PCD & EDM-PCD & PVD & GeM3D & Ours \\
                \midrule
                {FID ($\downarrow$)} & 95.83 & \underline{51.50} & 66.35 & 133.7 & \textbf{35.99}\\
                {KID ($\downarrow$)} & 10.99 & \underline{3.544} & 5.057 & 15.69 & \textbf{2.204}\\
                {MMD-CD ($\downarrow$)}  & \textbf{6.923} & 8.044 & 8.411 & 9.537 & \underline {7.720}\\
                {MMD-EMD ($\downarrow$)}  & \underline{0.941} & 1.002 & 1.049 & 1.095 & \textbf{0.930}\\
                {COV-CD ($\uparrow$)}  & \underline{70.60} & 58.06 & 48.16 & 38.11 & \textbf{72.87} \\
                {COV-EMD ($\uparrow$)}  & \underline{67.87} & 57.05 & 48.02 & 39.16 & \textbf{72.00}\\
                {1-NNA-CD ($\%, \downarrow$)}  & \underline{77.98} &  81.87 & 86.47 & 91.23 & \textbf{75.15}\\
                {1-NNA-EMD ($\%, \downarrow$)}  & \underline{79.57} & 81.09 & 86.27 & 89.72 & \textbf{76.47}\\
                \midrule
                {GTS (s, $\downarrow$)} & \textbf{1.32} & 6.46 & 3.94 & 55.67 & \underline{1.36} \\
                 \bottomrule
                \end{tabular} 
                }
\end{table}

%% file: tables/vessel.tex
\begin{table*}[!hbt]
\caption{Shape reconstruction and generation evaluation on two vessel datasets: CoW and ImageCAS. The best results are denoted in \textbf{Bold}.}
\label{tab:vessel}
        \centering
        \setlength\tabcolsep{4pt} 
        \footnotesize
     \resizebox{\textwidth}{!}{%
            \begin{tabular}{cc|cccc|cccccccc}
                \toprule
                \multirow{3}{*}{Dataset} & \multicolumn{1}{c}{\multirow{3}{*}{Method}} &
                 \multicolumn{4}{c}{Reconstruction} & 
                 \multicolumn{8}{c}{Generation} \\
                 \cmidrule(lr){3-6}
                 \cmidrule(lr){7-14}
                & \multicolumn{1}{c}{} & \multicolumn{1}{c}{\multirow{2}{*}{CD ($\downarrow$)}} &
                \multirow{2}{*}{EMD ($\downarrow$)} &
                \multirow{2}{*}{HD ($\downarrow$)} &
                \multicolumn{1}{c}{\multirow{2}{*}{F1 ($\%, \uparrow$)}} &
                \multirow{2}{*}{FID ($\downarrow$)}  & \multirow{2}{*}{KID ($\downarrow$)} & 
                 \multicolumn{2}{c}{MMD ($\downarrow$) } & 
                 \multicolumn{2}{c}{COV ($\%, \uparrow$) } &
                 \multicolumn{2}{c}{1-NNA ($\%, \downarrow$) }\\
                 \cmidrule(lr){9-10}
                 \cmidrule(lr){11-12}
                 \cmidrule(lr){13-14}
                & \multicolumn{1}{c}{}& & & & \multicolumn{1}{c}{} & & & CD & EMD & CD & EMD & CD & EMD\\
                \midrule
                \multirow{2}{*}{CoW} & Diff-Vessel & 4.907 & 1.041 & 2.502 & 71.79 &  23.08 & 1.422 & \textbf{4.978} & \textbf{0.733} & 67.27 & 52.72 & 91.17 & 91.35 \\
                & Ours & \textbf{1.796} & \textbf{0.321} & \textbf{0.840} & \textbf{99.78}  & \textbf{18.28} & \textbf{0.725} & 5.197 & 0.749 & \textbf{81.82} & \textbf{80.00} & \textbf{88.64} & \textbf{89.36}\\
                \midrule
                \multirow{2}{*}{ImageCAS} & Diff-Vessel & 5.810 & 1.364 & 3.485 & 65.42 &  34.26 & 1.804 & 8.471 & 1.130 & 59.50 & 54.50 & 84.14 & 82.14\\
                & Ours & \textbf{1.314} & \textbf{0.272} & \textbf{0.428} & \textbf{99.98}  & \textbf{24.18} & \textbf{1.432} & \textbf{7.377} & \textbf{0.975} & \textbf{76.00} & \textbf{66.50} & \textbf{64.85} & \textbf{64.28}\\
                 \bottomrule 
                \end{tabular} 
                }
\vspace{-1.0em}
\end{table*}

%% file: tables/ablation.tex
\begin{table*}[h]
\caption{Ablation study on the CoW, ImageCAS, and MedSDF dataset. The best results are denoted in \textbf{Bold}. Recommended settings are marked with grey.}
\label{tab:ablation}

        \centering
        \setlength\tabcolsep{4pt} 
        \footnotesize
        \resizebox{\textwidth}{!}{%
            \begin{tabular}{cccc|cccc|cccccccc}
                \toprule
                \multirow{3}{*}{Dataset} & \multicolumn{3}{c}{\multirow{1}{*}{Module}} &
                 \multicolumn{4}{c}{Reconstruction} & 
                 \multicolumn{8}{c}{Generation} \\
                 \cmidrule(lr){2-4}
                 \cmidrule(lr){5-8}
                 \cmidrule(lr){9-16}
                & \multicolumn{1}{c}{\multirow{2}{*}{DS}} & \multicolumn{1}{c}{\multirow{2}{*}{SC}} & \multicolumn{1}{c}{\multirow{2}{*}{LA}} & \multicolumn{1}{c}{\multirow{2}{*}{CD ($\downarrow$)}} &
                \multirow{2}{*}{EMD ($\downarrow$)} &
                \multirow{2}{*}{HD ($\downarrow$)} &
                \multicolumn{1}{c}{\multirow{2}{*}{F1 ($\%, \uparrow$)}} &
                \multirow{2}{*}{FID ($\downarrow$)}  & \multirow{2}{*}{KID ($\downarrow$)} & 
                 \multicolumn{2}{c}{MMD ($\downarrow$) } & 
                 \multicolumn{2}{c}{COV ($\%, \uparrow$) } &
                 \multicolumn{2}{c}{1-NNA ($\%, \downarrow$) }\\
                 \cmidrule(lr){11-12}
                 \cmidrule(lr){13-14}
                 \cmidrule(lr){15-16}
                & & & \multicolumn{1}{c}{}& & & & \multicolumn{1}{c}{} & & & CD & EMD & CD & EMD & CD & EMD\\
                \midrule
                \multirow{5}{*}{CoW} & \XSolidBrush & \XSolidBrush & \XSolidBrush & 4.102 & 0.822 & 1.847 & 80.39 &  34.77 & 4.106 & 5.249 & 0.741 & \textbf{81.82} & \textbf{80.00} & 92.61 & 91.89\\
                 & \CheckmarkBold & \XSolidBrush & \XSolidBrush & 2.348 & 0.481 & 1.293 & 98.13 &  23.57 & 1.844 & \textbf{4.911} & \textbf{0.725} & 76.36 & 72.72 & 91.53 & 91.35 \\
                 & \CheckmarkBold & \CheckmarkBold & \XSolidBrush & 2.277 & 0.487 & 1.222 & 98.23 &  22.63 & 1.472 & 4.988 & 0.734 & 76.36 & 74.54 & 91.89 & 90.99\\
                 & \CheckmarkBold & \CheckmarkBold & \CheckmarkBold & \textbf{1.796} & \textbf{0.321} & \textbf{0.839} & \textbf{99.78} &  \textbf{18.28} & \textbf{0.725} & 5.197 & 0.749 & \textbf{81.82} & \textbf{80.00} & \textbf{88.64} & \textbf{89.36}\\
                \midrule
                \multirow{5}{*}{ImageCAS} & \XSolidBrush & \XSolidBrush & \XSolidBrush & 1.443 & 0.358 & 0.531 & 99.92 &  \textbf{23.97} & 1.612 & 6.941 & 0.949 & 67.50 & 59.50 & 71.85 & 73.85 \\
                 & \CheckmarkBold & \XSolidBrush & \XSolidBrush & 1.460 & 0.328 & 0.485 & 99.98 &  25.36 & 1.643 & 6.793 & 0.937 & 67.50 & 59.50 & 72.28 & 72.57\\
                 & \CheckmarkBold & \CheckmarkBold & \XSolidBrush & 1.425 & 0.323 & 0.494 & \textbf{99.99} &  24.39 & 1.639 & \textbf{6.761} & \textbf{0.930} & 69.50 & 60.50 & 70.85 & 72.71\\
                 & \CheckmarkBold & \CheckmarkBold & \CheckmarkBold & \textbf{1.314} & \textbf{0.272} & \textbf{0.428} & 99.98 &  23.98 & \textbf{1.432} & 7.377 & 0.972 & \textbf{76.00} & \textbf{66.50} & \textbf{64.85} & \textbf{64.28} \\
                 \midrule
                \multirow{5}{*}{MedSDF} & \XSolidBrush & \XSolidBrush & \XSolidBrush & 3.716 & 1.031  & 2.580 & 93.25 &  85.34  & 8.115 & 8.250 & 1.014 & 68.35  &  68.28 & 78.72 & 77.25 \\
                 & \CheckmarkBold & \XSolidBrush & \XSolidBrush & 2.528 & 0.509 & 1.256 & 97.66 &  45.21 & 3.245 & 7.685 & 0.929 & 69.10 & 67.99 & 78.59 & 79.00 \\
                 & \CheckmarkBold & \CheckmarkBold & \XSolidBrush & 2.566 & 0.484 & 1.275 & 97.42 &  40.95 & 2.804 & \textbf{7.665} & \textbf{0.928} & 72.63 & 71.07 &76.65 & 77.32\\
                 & \CheckmarkBold & \CheckmarkBold & \CheckmarkBold & \textbf{2.314} & \textbf{0.447} & \textbf{1.136} & \textbf{98.24} &  \textbf{35.99} & \textbf{2.204} & 7.716 & 0.930 & \textbf{72.87} & \textbf{72.00} & \textbf{75.15} & \textbf{76.47}\\
                 \bottomrule 
                \end{tabular} 
                }
\vspace{-1.0em}
\end{table*}

%% file: sections/discussion.tex
\section{Discussion}
We propose a skeletal latent diffusion framework for medical shape generation. Our method first captures global geometry and topology through an online differentiable skeletonization module. Based on the extracted skeleton, we train a shape auto-encoder to learn compact skeletal latents and estimate neural implicit fields. Shape generation is achieved through a well-trained diffusion model in the latent space, followed by SDF decoding and mesh extraction. To support comprehensive evaluation, we construct MedSDF, a large-scale multi-category medical shape dataset containing surface point clouds and SDF volumes. Experiments on MedSDF and vascular datasets demonstrate that our method achieves superior reconstruction and generation quality while improving computational efficiency.

Despite these advantages, several limitations remain. The current geometric skeleton extraction may suffer from topological discontinuities. In addition, the constructed dataset focuses primarily on organ surfaces, whereas modeling internal anatomical structures is also important for medical analysis. In future work, we will explore learning-based skeletonization methods with topology-aware constraints to improve structural consistency. We also plan to expand the dataset with additional anatomical categories and more complex organs, and extend the framework toward large-scale modeling and analysis of the whole human body.

%% file: main.bbl
\begin{thebibliography}{42}
\providecommand{\natexlab}[1]{#1}
\providecommand{\url}[1]{#1}
\csname url@samestyle\endcsname
\providecommand{\newblock}{\relax}
\providecommand{\bibinfo}[2]{#2}
\providecommand{\BIBentrySTDinterwordspacing}{\spaceskip=0pt\relax}
\providecommand{\BIBentryALTinterwordstretchfactor}{4}
\providecommand{\BIBentryALTinterwordspacing}{\spaceskip=\fontdimen2\font plus
\BIBentryALTinterwordstretchfactor\fontdimen3\font minus \fontdimen4\font\relax}
\providecommand{\BIBforeignlanguage}[2]{{%
\expandafter\ifx\csname l@#1\endcsname\relax
\typeout{** WARNING: IEEEtranN.bst: No hyphenation pattern has been}%
\typeout{** loaded for the language `#1'. Using the pattern for}%
\typeout{** the default language instead.}%
\else
\language=\csname l@#1\endcsname
\fi
#2}}
\providecommand{\BIBdecl}{\relax}
\BIBdecl

\bibitem[Luo and Hu(2021)]{luo2021diffusion}
S.~Luo and W.~Hu, ``Diffusion probabilistic models for 3d point cloud generation,'' in \emph{Proceedings of the IEEE/CVF Conference on Computer Vision and Pattern Recognition (CVPR)}, June 2021.

\bibitem[Zhou et~al.(2021)Zhou, Du, and Wu]{zhou20213d}
L.~Zhou, Y.~Du, and J.~Wu, ``3d shape generation and completion through point-voxel diffusion,'' in \emph{Proceedings of the IEEE/CVF international conference on computer vision}, 2021, pp. 5826--5835.

\bibitem[Zeng et~al.(2022)Zeng, Vahdat, Williams, Gojcic, Litany, Fidler, and Kreis]{zeng2022lion}
X.~Zeng, A.~Vahdat, F.~Williams, Z.~Gojcic, O.~Litany, S.~Fidler, and K.~Kreis, ``Lion: Latent point diffusion models for 3d shape generation,'' in \emph{Advances in Neural Information Processing Systems (NeurIPS)}, 2022.

\bibitem[Sinha and Hamarneh(2024)]{Sin_TrIND_MICCAI2024}
A.~Sinha and G.~Hamarneh, ``{ TrIND: Representing Anatomical Trees by Denoising Diffusion of Implicit Neural Fields },'' in \emph{proceedings of Medical Image Computing and Computer Assisted Intervention -- MICCAI 2024}, vol. LNCS 15012.\hskip 1em plus 0.5em minus 0.4em\relax Springer Nature Switzerland, October 2024.

\bibitem[Prabhakar et~al.(2024)Prabhakar, Shit, Musio, Yang, Amiranashvili, Paetzold, Li, and Menze]{Pra_3D_MICCAI2024}
C.~Prabhakar, S.~Shit, F.~Musio, K.~Yang, T.~Amiranashvili, J.~C. Paetzold, H.~B. Li, and B.~Menze, ``{ 3D Vessel Graph Generation Using Denoising Diffusion },'' in \emph{proceedings of Medical Image Computing and Computer Assisted Intervention -- MICCAI 2024}, vol. LNCS 15011.\hskip 1em plus 0.5em minus 0.4em\relax Springer Nature Switzerland, October 2024.

\bibitem[Chen et~al.(2025)Chen, Zhang, Lai, Shen, Zhang, Dong, Chen, and Li]{CheSiq_Hierarchical_MICCAI2025}
S.~Chen, G.~Zhang, J.~Lai, B.~Shen, S.~Zhang, C.~Dong, X.~Chen, and Y.~Li, ``{ Hierarchical Part-based Generative Model for Realistic 3D Blood Vessel },'' in \emph{proceedings of Medical Image Computing and Computer Assisted Intervention -- MICCAI 2025}, vol. LNCS 15962.\hskip 1em plus 0.5em minus 0.4em\relax Springer Nature Switzerland, September 2025.

\bibitem[Guo et~al.(2025)Guo, Tan, Feng, and Zhou]{guo2025vesseldiffusion}
Z.~Guo, Z.~Tan, J.~Feng, and J.~Zhou, ``Vesseldiffusion: 3d vascular structure generation based on diffusion model,'' \emph{IEEE Transactions on Medical Imaging}, 2025.

\bibitem[Petrov et~al.(2024)Petrov, Goyal, Thamizharasan, Kim, Gadelha, Averkiou, Chaudhuri, and Kalogerakis]{petrov2024gem3d}
D.~Petrov, P.~Goyal, V.~Thamizharasan, V.~Kim, M.~Gadelha, M.~Averkiou, S.~Chaudhuri, and E.~Kalogerakis, ``Gem3d: Generative medial abstractions for 3d shape synthesis,'' in \emph{ACM SIGGRAPH 2024 Conference Papers}, 2024, pp. 1--11.

\bibitem[Karras et~al.(2022)Karras, Aittala, Aila, and Laine]{Karras2022edm}
T.~Karras, M.~Aittala, T.~Aila, and S.~Laine, ``Elucidating the design space of diffusion-based generative models,'' in \emph{Proc. NeurIPS}, 2022.

\bibitem[Amenta et~al.(1998)Amenta, Bern, and Kamvysselis]{10.1145/280814.280947}
N.~Amenta, M.~Bern, and M.~Kamvysselis, ``A new voronoi-based surface reconstruction algorithm,'' in \emph{Proceedings of the 25th Annual Conference on Computer Graphics and Interactive Techniques}, ser. SIGGRAPH '98.\hskip 1em plus 0.5em minus 0.4em\relax New York, NY, USA: Association for Computing Machinery, 1998, p. 415–421.

\bibitem[Yan et~al.(2018)Yan, Letscher, and Ju]{10.1145/3197517.3201396}
Y.~Yan, D.~Letscher, and T.~Ju, ``Voxel cores: efficient, robust, and provably good approximation of 3d medial axes,'' \emph{ACM Trans. Graph.}, vol.~37, no.~4, Jul. 2018.

\bibitem[Wang et~al.(2022)Wang, Wang, Wang, and Guo]{10.1145/3550454.3555465}
N.~Wang, B.~Wang, W.~Wang, and X.~Guo, ``Computing medial axis transform with feature preservation via restricted power diagram,'' \emph{ACM Trans. Graph.}, vol.~41, no.~6, Nov. 2022.

\bibitem[Huang et~al.(2013)Huang, Wu, Cohen-Or, Gong, Zhang, Li, and Chen]{10.1145/2461912.2461913}
H.~Huang, S.~Wu, D.~Cohen-Or, M.~Gong, H.~Zhang, G.~Li, and B.~Chen, ``L1-medial skeleton of point cloud,'' \emph{ACM Trans. Graph.}, vol.~32, no.~4, Jul. 2013.

\bibitem[Tagliasacchi et~al.(2012)Tagliasacchi, Alhashim, Olson, and Zhang]{10.1111/j.1467-8659.2012.03178.x}
A.~Tagliasacchi, I.~Alhashim, M.~Olson, and H.~Zhang, ``Mean curvature skeletons,'' \emph{Comput. Graph. Forum}, vol.~31, no.~5, p. 1735–1744, Aug. 2012.

\bibitem[Tagliasacchi et~al.(2016)Tagliasacchi, Delame, Spagnuolo, Amenta, and Telea]{tagliasacchi20163d}
A.~Tagliasacchi, T.~Delame, M.~Spagnuolo, N.~Amenta, and A.~Telea, ``3d skeletons: A state-of-the-art report,'' in \emph{Computer Graphics Forum}, vol.~35, no.~2.\hskip 1em plus 0.5em minus 0.4em\relax Wiley Online Library, 2016, pp. 573--597.

\bibitem[Yin et~al.(2018)Yin, Huang, Cohen-Or, and Zhang]{yin2018p2pnet}
K.~Yin, H.~Huang, D.~Cohen-Or, and H.~Zhang, ``P2p-net: Bidirectional point displacement net for shape transform,'' \emph{ACM Transactions on Graphics(Special Issue of SIGGRAPH)}, vol.~37, no.~4, pp. 152:1--152:13, 2018.

\bibitem[Ge et~al.(2023)Ge, Yao, Yang, Wang, Chen, and Guo]{10.1016/j.cag.2023.07.020}
M.~Ge, J.~Yao, B.~Yang, N.~Wang, Z.~Chen, and X.~Guo, ``Point2mm: Learning medial mesh from point clouds,'' \emph{Comput. Graph.}, vol. 115, no.~C, p. 511–521, Oct. 2023.

\bibitem[Lin et~al.(2021)Lin, Li, Liu, Chen, Choi, and Wang]{Lin_2021_CVPR}
C.~Lin, C.~Li, Y.~Liu, N.~Chen, Y.-K. Choi, and W.~Wang, ``Point2skeleton: Learning skeletal representations from point clouds,'' in \emph{Proceedings of the IEEE/CVF Conference on Computer Vision and Pattern Recognition (CVPR)}, June 2021, pp. 4277--4286.

\bibitem[Ho et~al.(2020)Ho, Jain, and Abbeel]{10.5555/3495724.3496298}
J.~Ho, A.~Jain, and P.~Abbeel, ``Denoising diffusion probabilistic models,'' in \emph{Proceedings of the 34th International Conference on Neural Information Processing Systems}, ser. NIPS '20.\hskip 1em plus 0.5em minus 0.4em\relax Red Hook, NY, USA: Curran Associates Inc., 2020.

\bibitem[Park et~al.(2019)Park, Florence, Straub, Newcombe, and Lovegrove]{Park_2019_CVPR}
J.~J. Park, P.~Florence, J.~Straub, R.~Newcombe, and S.~Lovegrove, ``Deepsdf: Learning continuous signed distance functions for shape representation,'' in \emph{The IEEE Conference on Computer Vision and Pattern Recognition (CVPR)}, June 2019.

\bibitem[Achlioptas et~al.(2018)Achlioptas, Diamanti, Mitliagkas, and Guibas]{pmlr-v80-achlioptas18a}
P.~Achlioptas, O.~Diamanti, I.~Mitliagkas, and L.~Guibas, ``Learning representations and generative models for 3{D} point clouds,'' in \emph{Proceedings of the 35th International Conference on Machine Learning}, ser. Proceedings of Machine Learning Research, J.~Dy and A.~Krause, Eds., vol.~80.\hskip 1em plus 0.5em minus 0.4em\relax PMLR, 10--15 Jul 2018, pp. 40--49.

\bibitem[Cheng et~al.(2022)Cheng, Lee, Tuyakov, Schwing, and Gui]{cheng2022sdfusion}
Y.-C. Cheng, H.-Y. Lee, S.~Tuyakov, A.~Schwing, and L.~Gui, ``{SDFusion}: Multimodal 3d shape completion, reconstruction, and generation,'' \emph{arXiv}, 2022.

\bibitem[Hu et~al.(2024)Hu, Hui, Liu, Li, and Fu]{10.1145/3635304}
J.~Hu, K.-H. Hui, Z.~Liu, R.~Li, and C.-W. Fu, ``Neural wavelet-domain diffusion for 3d shape generation, inversion, and manipulation,'' \emph{ACM Trans. Graph.}, vol.~43, no.~2, Jan. 2024.

\bibitem[Zhang et~al.(2023)Zhang, Tang, Nie\ss{}ner, and Wonka]{10.1145/3592442}
B.~Zhang, J.~Tang, M.~Nie\ss{}ner, and P.~Wonka, ``3dshape2vecset: A 3d shape representation for neural fields and generative diffusion models,'' \emph{ACM Trans. Graph.}, vol.~42, no.~4, jul 2023.

\bibitem[Lorensen and Cline(1987)]{10.1145/37401.37422}
W.~E. Lorensen and H.~E. Cline, ``Marching cubes: A high resolution 3d surface construction algorithm,'' in \emph{Proceedings of the 14th Annual Conference on Computer Graphics and Interactive Techniques}, ser. SIGGRAPH '87.\hskip 1em plus 0.5em minus 0.4em\relax New York, NY, USA: Association for Computing Machinery, 1987, p. 163–169.

\bibitem[Zhang and Wonka(2025)]{zhang2025lagem}
B.~Zhang and P.~Wonka, ``{LaGeM}: A large geometry model for 3d representation learning and diffusion,'' in \emph{The Thirteenth International Conference on Learning Representations}, 2025.

\bibitem[Ester et~al.(1996)Ester, Kriegel, Sander, and Xu]{10.5555/3001460.3001507}
M.~Ester, H.-P. Kriegel, J.~Sander, and X.~Xu, ``A density-based algorithm for discovering clusters in large spatial databases with noise,'' in \emph{Proceedings of the Second International Conference on Knowledge Discovery and Data Mining}, ser. KDD'96.\hskip 1em plus 0.5em minus 0.4em\relax AAAI Press, 1996, p. 226–231.

\bibitem[Qi et~al.(2017{\natexlab{a}})Qi, Su, Mo, and Guibas]{qi2017pointnet}
C.~R. Qi, H.~Su, K.~Mo, and L.~J. Guibas, ``Pointnet: Deep learning on point sets for 3d classification and segmentation,'' in \emph{Proceedings of the IEEE conference on computer vision and pattern recognition}, 2017, pp. 652--660.

\bibitem[Qi et~al.(2017{\natexlab{b}})Qi, Yi, Su, and Guibas]{qi2017pointnet++}
C.~R. Qi, L.~Yi, H.~Su, and L.~J. Guibas, ``Pointnet++: Deep hierarchical feature learning on point sets in a metric space,'' \emph{Advances in neural information processing systems}, vol.~30, 2017.

\bibitem[Vaswani et~al.(2017)Vaswani, Shazeer, Parmar, Uszkoreit, Jones, Gomez, Kaiser, and Polosukhin]{vaswani2017attention}
A.~Vaswani, N.~Shazeer, N.~Parmar, J.~Uszkoreit, L.~Jones, A.~N. Gomez, {\L}.~Kaiser, and I.~Polosukhin, ``Attention is all you need,'' \emph{Advances in neural information processing systems}, vol.~30, 2017.

\bibitem[Newcombe et~al.(2011)Newcombe, Izadi, Hilliges, Molyneaux, Kim, Davison, Kohi, Shotton, Hodges, and Fitzgibbon]{6162880}
R.~A. Newcombe, S.~Izadi, O.~Hilliges, D.~Molyneaux, D.~Kim, A.~J. Davison, P.~Kohi, J.~Shotton, S.~Hodges, and A.~Fitzgibbon, ``Kinectfusion: Real-time dense surface mapping and tracking,'' in \emph{2011 10th IEEE International Symposium on Mixed and Augmented Reality}, 2011, pp. 127--136.

\bibitem[Rombach et~al.(2021)Rombach, Blattmann, Lorenz, Esser, and Ommer]{rombach2021highresolution}
R.~Rombach, A.~Blattmann, D.~Lorenz, P.~Esser, and B.~Ommer, ``High-resolution image synthesis with latent diffusion models,'' 2021.

\bibitem[Ho and Salimans(2021)]{ho2021classifierfree}
J.~Ho and T.~Salimans, ``Classifier-free diffusion guidance,'' in \emph{NeurIPS 2021 Workshop on Deep Generative Models and Downstream Applications}, 2021.

\bibitem[Li et~al.(2024)Li, Zhou, Yang, Pepe, Gsaxner, Luijten, Qu, Zhang, Chen, Li, et~al.]{li2024medshapenet}
J.~Li, Z.~Zhou, J.~Yang, A.~Pepe, C.~Gsaxner, G.~Luijten, C.~Qu, T.~Zhang, X.~Chen, W.~Li \emph{et~al.}, ``Medshapenet--a large-scale dataset of 3d medical shapes for computer vision,'' \emph{Biomedical Engineering/Biomedizinische Technik}, no.~0, 2024.

\bibitem[Zeng et~al.(2023)Zeng, Wu, Lin, Xie, Hong, Huang, Zhuang, Bi, Pan, Ullah, Khan, Wang, Shi, Li, and Xu]{ZENG2023102287}
A.~Zeng, C.~Wu, G.~Lin, W.~Xie, J.~Hong, M.~Huang, J.~Zhuang, S.~Bi, D.~Pan, N.~Ullah, K.~N. Khan, T.~Wang, Y.~Shi, X.~Li, and X.~Xu, ``Imagecas: A large-scale dataset and benchmark for coronary artery segmentation based on computed tomography angiography images,'' \emph{Computerized Medical Imaging and Graphics}, vol. 109, p. 102287, 2023.

\bibitem[Junayed et~al.(2024)Junayed, Sanjid, Hossain, Uddin, and Haque]{Junayed_2024}
M.~S.~S. Junayed, K.~S. Sanjid, M.~T. Hossain, M.~M. Uddin, and S.~A. Haque, ``Topology‐aware anatomical segmentation of the circle of willis: Hunet unveils the vascular network,'' \emph{IET Image Processing}, vol.~18, no.~10, p. 2745–2753, Jun. 2024.

\bibitem[Wang et~al.(2019)Wang, Sun, Liu, Sarma, Bronstein, and Solomon]{wang2019dynamic}
Y.~Wang, Y.~Sun, Z.~Liu, S.~E. Sarma, M.~M. Bronstein, and J.~M. Solomon, ``Dynamic graph cnn for learning on point clouds,'' \emph{ACM Transactions on Graphics (tog)}, vol.~38, no.~5, pp. 1--12, 2019.

\bibitem[Zhang et~al.(2024)Zhang, Yang, and Li]{zhang2024flemme}
G.~Zhang, J.~Yang, and Y.~Li, ``Flemme: A flexible and modular learning platform for medical images,'' in \emph{2024 IEEE International Conference on Bioinformatics and Biomedicine (BIBM)}.\hskip 1em plus 0.5em minus 0.4em\relax IEEE, 2024, pp. 4018--4023.

\bibitem[Guo et~al.(2021)Guo, Cai, Liu, Mu, Martin, and Hu]{guo2021pct}
M.-H. Guo, J.-X. Cai, Z.-N. Liu, T.-J. Mu, R.~R. Martin, and S.-M. Hu, ``Pct: Point cloud transformer,'' \emph{Computational Visual Media}, vol.~7, pp. 187--199, 2021.

\bibitem[Zhang et~al.(2025)Zhang, Yang, and Li]{ZhaGuo_Hierarchical_MICCAI2025}
G.~Zhang, J.~Yang, and Y.~Li, ``{ Hierarchical Feature Learning for Medical Point Clouds via State Space Model },'' in \emph{proceedings of Medical Image Computing and Computer Assisted Intervention -- MICCAI 2025}, vol. LNCS 15969.\hskip 1em plus 0.5em minus 0.4em\relax Springer Nature Switzerland, September 2025.

\bibitem[Yang et~al.(2018)Yang, Feng, Shen, and Tian]{yang2018foldingnet}
Y.~Yang, C.~Feng, Y.~Shen, and D.~Tian, ``Foldingnet: Point cloud auto-encoder via deep grid deformation,'' in \emph{Proceedings of the IEEE conference on computer vision and pattern recognition}, 2018, pp. 206--215.

\bibitem[Yang et~al.(2019)Yang, Huang, Hao, Liu, Belongie, and Hariharan]{9010395}
G.~Yang, X.~Huang, Z.~Hao, M.-Y. Liu, S.~Belongie, and B.~Hariharan, ``Pointflow: 3d point cloud generation with continuous normalizing flows,'' in \emph{2019 IEEE/CVF International Conference on Computer Vision (ICCV)}, 2019, pp. 4540--4549.

\end{thebibliography}
